\pdfoutput=1

\documentclass[runningheads]{llncs}
\usepackage{graphicx}

\usepackage{tikz}
\usepackage{comment}
\usepackage{amsmath,amssymb} 
\usepackage{color}
\usepackage{orcidlink}

\usepackage[accsupp]{axessibility}  
\usepackage{multirow}
\usepackage[normalem]{ulem}

\usepackage{multirow}

\newcommand{\mcl}[1]{\mathcal{#1}}

\newcommand{\norm}[1]{\left\|#1\right\|}

\newcommand{\eqn}[1]{\begin{equation}#1\end{equation}}
\newcommand{\eqna}[1]{\begin{eqnarray}#1\end{eqnarray}}

\newcommand{\itm}[1]{\begin{itemize}#1\end{itemize}}
\newcommand{\prn}[1]{\left(#1\right)}

\newcommand{\x}{\times}

\newcommand{\RR}{\mathbb{R}}

\newcommand{\tf}[1]{\textbf{#1}}

\begin{document}
\pagestyle{headings}
\mainmatter
\def\ECCVSubNumber{3747}  

\title{Learning Implicit Templates for Point-Based Clothed Human Modeling} 

\titlerunning{Learning Implicit Templates for Point-Based Clothed Human Modeling}
%
\author{Siyou Lin\orcidlink{0000-0002-8906-657X} \and
Hongwen Zhang\orcidlink{0000-0001-8633-4551} \and
Zerong Zheng\orcidlink{0000-0003-1339-2480} \and
Ruizhi Shao\orcidlink{0000-0003-2188-1348} \and
Yebin Liu\orcidlink{0000-0003-3215-0225}}
%
\authorrunning{S. Lin et al.}

\institute{Tsinghua University, Beijing, China}
\maketitle

\begin{abstract}
We present FITE, a First-Implicit-Then-Explicit framework for modeling human avatars in clothing. Our framework first learns implicit surface templates representing the coarse clothing topology, and then employs the templates to guide the generation of point sets which further capture pose-dependent clothing deformations such as wrinkles. Our pipeline incorporates the merits of both implicit and explicit representations, namely, the ability to handle varying topology and the ability to efficiently capture fine details. We also propose diffused skinning to facilitate template training especially for loose clothing, and projection-based pose-encoding to extract pose information from mesh templates without predefined UV map or connectivity. Our code is publicly available at \href{https://github.com/jsnln/fite}{https://github.com/jsnln/fite}.

\end{abstract}

\section{Introduction}

The modeling of clothed human avatars is an important topic in many graphics-related fields, such as animation, video games, virtual reality, etc. Traditional solutions~\cite{baran2007pinocchio,feng2015avatar,liu2019neuroskinning,loper2015smpl} are mostly based on rigging and skinning artist-designed avatars, and thus lack realism in representing clothed humans. Another possibility is to apply physics-based simulation~\cite{deformdynamics,guan2012drape,gundogdu2019garnet,patel20tailornet} which is in general computationally heavy and requires manually designed outfits. Recent work explores learning realistic pose-dependent deformations (e.g. wrinkles) of clothes from posed scan data. From the data-driven perspective, we identify two major challenges in this task: (i) learning the clothing topology; (ii) learning fine details and pose-dependent clothing deformations. In fact, most methods are limited by the representation they choose and cannot fully tackle the challenges above. In particular, mesh-based methods~\cite{alldieck2019learning,alldieck2019tex2shape,bhatnagar2019mgn,burov2021dsfn,de2010stable,corona2021smplicit,guan2012drape,gundogdu2019garnet,lahner2018deepwrinkles,jiang2020bcnet,ma2020cape,Neophytou2014layered,patel20tailornet,santesteban2019,tiwari20sizer,vidaurre2020fully,yang2018physics} are essentially limited by the fixed topology and typically require registered scans for training. On the other hand, implicit surfaces~\cite{chen2021snarf,chibane2020ndf,deng2019neural,gropp2020igr,mihajlovic2021leap,mescheder2019occupancy,park2019deepsdf,saito2020pifuhd,saito2019pifu,saito2021scanimate,wang2021metaavatar} can represent varying topology, but are computationally heavy and struggle to represent details. Point sets, on the other hand, enjoys efficiency as well as flexibility. However, generating point sets with details is difficult. Most methods generate sparse points~\cite{achlioptas2017learning,fan2017psg,lin2018learning} or points grouped into patches~\cite{bednarik2020,deng2020better,deprelle2019learning,groueix2018atlasnet,ma2021scale}. Although some methods achieve higher quality, they typically require dozens of iterations for a single output~\cite{klokov2020discrete,luo2021diffusion,Yang2019pointflow}.

Recently, the state-of-the-art (SOTA) point-based method, POP \cite{ma2021pop} demonstrates the power of points for cross-outfit modeling and for capturing pose-dependent clothing details. The success of POP lies in its robust body-template-plus-offsets formulation, the flexibility of point sets, and its fine-grained UV-space features. However, POP applies the same minimal body~\cite{loper2015smpl,pavlakos2019smplx} for all outfits, which suffers from artifacts such as overly sparse points and discontinuity in clothing, negatively affecting the overall visual quality.

The analysis above suggests implicit surfaces and point sets are complementary in a way that: (i) implicit surfaces can handle varying topology but do not efficiently converge to details; (ii) point sets are efficient and can represent fine details, but even the SOTA point-based scheme \cite{ma2021pop} is limited to an underlying template with fixed topology, leading to topology-related artifacts. This makes us wonder: Can we incorporate the merits of both types of representations to simultaneously capture the overall topology and final details? With this as motivation, we propose a \tf{First-Implicit-Then-Explicit} framework, abbreviated \tf{FITE}, where the implicit representation and the point set representation are tasked to do what they excel at. Our proposal is a two-stage pipeline: In stage one we train implicit templates that capture the coarse clothing topology for each outfit, and in stage two we predict pose-dependent offsets from the template to generate fine details and pose-dependent clothing deformations. To avoid any conceptual confusion with related work~\cite{genova2019sif,zheng2021dit}, we define an {\it implicit template} as a canonically posed clothed body associated with linear blend skinning weights. Since the templates already capture the coarse topology of given outfits, the second stage can focus on pose-dependent deformations. Compared with POP \cite{ma2021pop} which directly employs a fixed body template for all outfits, our divide-and-conquer scheme leads to better topology as well as better details. Note that our templates resemble the canonical-space shapes in \cite{chen2021snarf,saito2021scanimate}, but pose-dependent deformation for the templates is not required in our setting. 

Two problems naturally arise with the formulation above. First, in stage one, the training of implicit templates from posed scans requires known correspondences between the canonical space and the posed spaces. Existing approaches~\cite{chen2021snarf,saito2021scanimate} learn such correspondences by predicting 3D skinning fields. However, they are less accurate in regions far away from the skeleton. To tackle this problem, we precompute a 3D skinning field by smoothly diffusing the skinning weights of SMPL~\cite{loper2015smpl} into the whole space, and fix it for subsequent training. This approach effectively reduces the number of learnable parameters and induces more stable correspondences, especially in the case of limited data. Additionally, since the skinning is diffused smoothly, it can handle loose clothing as well. Another problem lies in stage two: How do we encode pose information for learned templates, which do not come with predefined UV or mesh connectivity? We propose to render the canonically posed template to multi-view images whose pixels are the coordinates of the corresponding posed vertices (following \cite{ma2021pop}, we refer to them as {\it position maps}), and feed them to U-Nets~\cite{ronneberger2015unet} to encode pose information. Compared with UV-space position maps~\cite{ma2021scale,ma2021pop}, our solution introduces a more continuous feature space for the templates and exhibits less topological artifacts.

We summarize our contributions as follows.
\itm{
    \item We propose a \tf{First-Implicit-Then-Explicit} framework for clothed human modeling which incorporates the merits of both implicit and explicit representations, and exhibits better topology properties than current methods.
    \item For coarse template training, we propose \tf{diffused skinning} which induces stable correspondences from the canonical space to posed spaces, even for limited data or loose clothing.
    \item For extracting pose information, we propose \tf{projection-based pose encoding} which introduces a continuous feature space on trained templates without predefined UV map or connectivity.
}

\section{Related Work}

Modeling clothed human avatars is a task involving various techniques. An ideal approach should at least (i) adopt a suitable 3D representation; (ii) be compatible with existing animation pipelines; (iii) model fine details and pose-dependent clothing deformations. We mainly focus on these three aspects in this review.

\subsection{Representations for Clothed Humans}

Mesh surfaces are to-date the predominant choice for representing 3D shapes for their compactness and efficiency. For human modeling, most methods represent clothing as deformations~\cite{alldieck2019tex2shape,bhatnagar2019mgn,burov2021dsfn,ma2020cape,Neophytou2014layered,su2022deepcloth,tiwari20sizer,yang2018physics} from minimal bodies~\cite{anguelov2005scape,hirshberg2012blendscape,loper2015smpl,pavlakos2019smplx}, or as separate layers~\cite{guan2012drape,gundogdu2019garnet,lahner2018deepwrinkles,patel20tailornet,santesteban2019}. Meshes are essentially limited their fixed topology. On the other hand, neural implicit human representations are not subject to a fixed topology~\cite{chen2021snarf,chibane2020ndf,deng2019neural,gropp2020igr,mihajlovic2021leap,mescheder2019occupancy,park2019deepsdf,saito2020pifuhd,saito2019pifu,saito2021scanimate,wang2021metaavatar}. Despite the obvious advantages, the low computational efficiency often forbids high-fidelity outputs by implicit methods. Articulating implicitly represented humans is also challenging. Recent methods~\cite{chen2021snarf,deng2020nasa,mihajlovic2021leap,saito2021scanimate} learn volumetric linear blend skinning to achieve articulation. However, extending linear blend skinning to 3D can be tricky, especially with limited data or loose clothing. Point sets enjoy efficiency as well as flexibility. Nonetheless, producing points with fine details is difficult. Pioneer work~\cite{achlioptas2017learning,fan2017psg,lin2018learning} only generates sparse points. Later approaches~\cite{bednarik2020,deng2020better,deprelle2019learning,groueix2018atlasnet,ma2021scale} achieve denser generation by grouping points into patches, with notable inter-patch discontinuity as a side-effect. POP~\cite{ma2021pop} conditions the generation on fine-grained UV features to produce dense structured points, but its performance it still limited by the underlying topology of SMPL/SMPL-X~\cite{loper2015smpl,pavlakos2019smplx}. Recently, based on neural radiance fields (NeRF)~\cite{mildenhall2021nerf}, attempts have been made to bypass the underlying geometry and synthesize rendered images of clothed humans directly~\cite{peng2021animatablenerf,peng2021neural,weng2022humannerf,zheng2022structured}. However, the lack of an explicit geometry limits their application in downstream tasks such as editing and animation.

\subsection{Animating Humans with Linear Blend Skinning}

Linear blend skinning (LBS) is a widely used technique for animating human avatars among other articulatable objects. With LBS, a surface is associated with an underlying skeleton, and when the skeleton is articulated, each point on the surface is also transformed by linearly combining the transformations of the bones in the skeleton, according to a set of predefined skinning weights.

LBS is traditionally only applied to 2D mesh surfaces~\cite{baran2007pinocchio,loper2015smpl,pavlakos2019smplx}. Recently, motivated by the need to articulate implicit representations, LBS has also been extended to 3D (called skinning fields)~\cite{bhatnagar2020loopreg,chen2021snarf,deng2020nasa,mihajlovic2021leap,saito2021scanimate}. However, there is no direct supervision on the skinning weights for locations far away from the skeleton. LoopReg~\cite{bhatnagar2020loopreg}
uses the nearest point on SMPL to extend skinning to 3D, leading to clearly observable spatial discontinuity. SNARF~\cite{chen2021snarf} trains a forward skinning field jointly with a canonical-space implicit occupancy field. SCANimate~\cite{saito2021scanimate} predicts both forward and backward skinning fields and enforce cycle consistency. Despite these efforts, such skinning fields are still limited to tight clothing.

\subsection{Modeling Pose-Dependent Clothing Deformations}

Clothing deformations are in general non-rigid and pose-dependent, e.g., wrinkles, sliding motions and bulging. Generating realistic pose-dependent deformations requires effectively encoding pose information. Going beyond prior methods that apply a single global feature~\cite{deng2020nasa,lahner2018deepwrinkles,ma2020cape,patel20tailornet,yang2018analyzing}, recent work has demonstrated utilizing local information leads to better details and better generalization~\cite{ma2021scale,ma2021pop,saito2021scanimate,su2022deepcloth}, e.g. UV maps~\cite{ma2021scale,ma2021pop,su2022deepcloth} and attention mechanisms~\cite{saito2021scanimate}.

In our framework, we propose projection-based pose encoding to introduce a continuous pose feature space on templates without predefined UV map or mesh connectivity. Although similar ideas have also appeared elsewhere~\cite{chan2021eg3d,peng2020con,saito2019pifu,saito2020pifuhd}, to the best of our knowledge, we are the first to apply such architectures to encode pose information for animating clothed humans. 

\section{Method}

\subsection{Task Formulation and Notations}

\begin{figure}[t]
\centering
\includegraphics[width=0.95\linewidth]{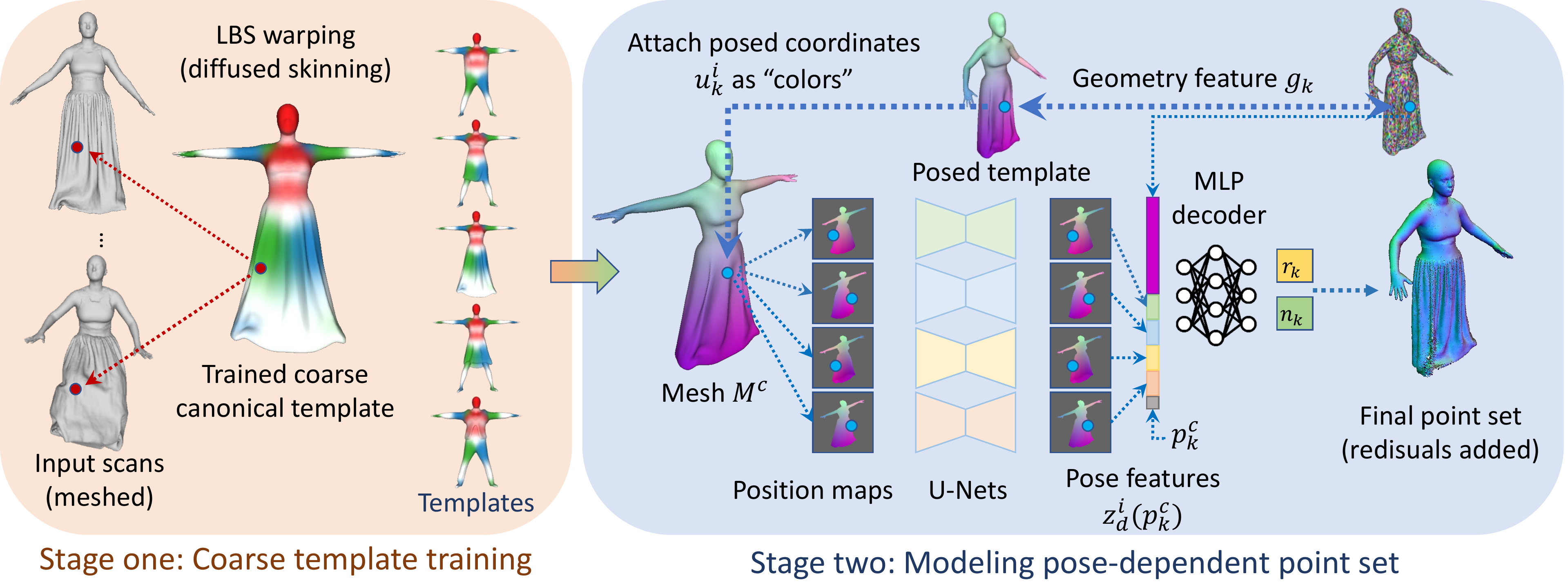}
\caption{Overall pipeline of our first-implicit-then-explicit framework. Left: In stage one we learn implicit templates of different outfits with diffused skinning. Right: In stage two we predict pose-dependent offset from features extracted by projection-based encoders.}
\label{fig:pipeline}
\end{figure}

Our task is to learn animatable clothed human avatars with realistic pose-dependent clothing deformations from a set of posed scans, under a multi-outfit setting. Fig.~\ref{fig:pipeline} shows our overall pipeline, where implicit templates are trained in stage one and pose-dependent offsets are predicted in stage two. For simplicity in notations let us for now assume a single outfit worn by the same person. We will introduce how the formulation can be easily extended to the multi-outfit setting at the end of Section~\ref{sec:stage2}.

We assume input scans are presented in the form of point sets with normals that cover most of the body so that watertight meshes can be extracted for obtaining ground truth occupancy labels ($0$ for outside and $1$ for inside). We denote the point set of the $i$-th posed scan as $\{p^i_k\}_{k=1}^{N_i}\subset\RR^3$, where $N_i$ is the number of points in the $i$-th scan. The normal at $p^i_k$ is denoted $n^i_k$. We also assume each scan has a fitted SMPL model \cite{loper2015smpl}. This assumption is reasonable since there are a number of existing techniques that can tractably obtain parametric body models~\cite{kanazawa17hmr,pavlakos2022multishot,tian2022recovering,zhang2021lightweight}. Note that SMPL is needed only for its skeletal structure and skinning weights for reposing, and that for point set generation our learned implicit templates will be used instead. Let $T$ denote the canonical-pose SMPL body template of the subject and let $\theta^i\in\RR^{72}$ denote the SMPL pose parameter corresponding to the $i$-th scan. $T$, together with $\theta^i$, determines a set of rigid transformations $R^i_j$ associated to each joint of the SMPL skeleton ($j=1,\cdots,24$ is the index for different joints).

Given $T$ and $\theta^i$, if a point $p\in\RR^3$ in the canonical space has skinning weights $w(p)=(w_1(p),\cdots,w_{24}(p))\in\RR^{24}$ associated to each joint, we can warp $p$ to its corresponding position $q^i$ in the $i$-th scan via LBS. We denote this warping as $W(\ \cdot\ ,\ \cdot\ ;T,\theta^i):\RR^3\x\RR^{24}\to\RR^3$. More specifically:
\eqn{
    q^i=W(p,w(p),T,\theta^i)=\sum_{j=1}^{24}w_j(p)R^i_j(p).\label{eq:warpingeq}
}
Note that $w(p)$ is for now only defined on the SMPL surface $T$. We introduce how to extend $w$ to the 3D space in Section~\ref{sec:diffskinning}.

\subsection{Stage One: Coarse Template Training with Diffused Skinning}\label{sec:diffskinning}

\begin{figure}
\centering
\includegraphics[width=0.95\linewidth]{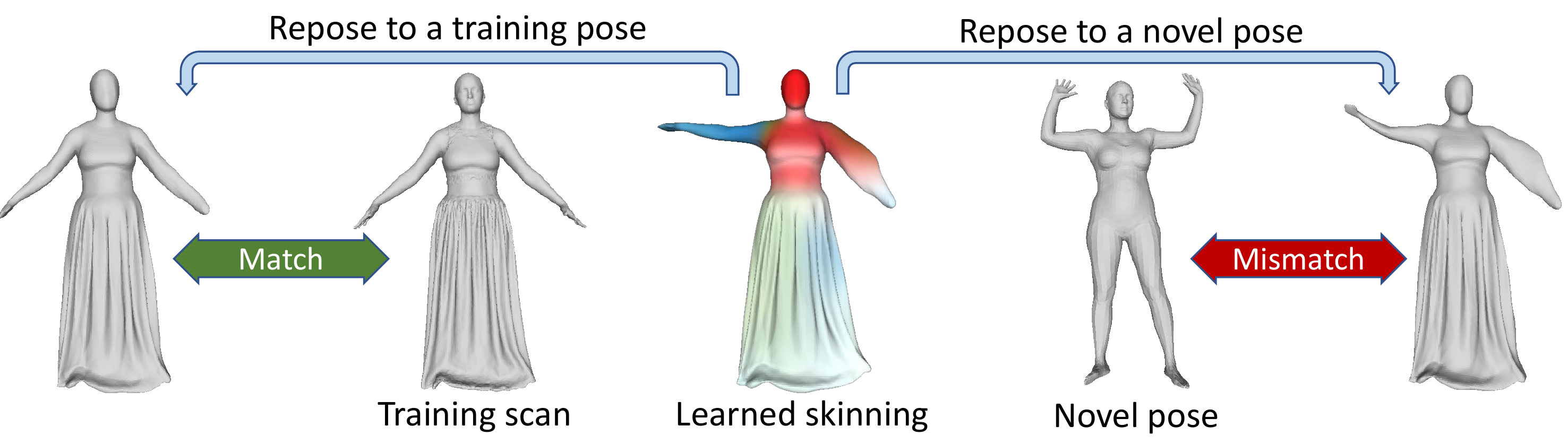}
\caption{Illustration of the ill-posedness of jointly optimizing the canonical shape and the skinning fields. An incorrect canonical shape with incorrect skinning can be accidentally warped to a correct pose, but generalization to new poses can be problematic.}
\label{fig:skinning_test}
\end{figure}

In this stage, we seek to obtain a template $T^c$ representing the coarse clothing topology. We follow SNARF~\cite{chen2021snarf} to learn the template as the $1/2$-level-set of a 0-1 occupancy field $F^c:\RR^3\to[0,1]$ in the canonical space:
\eqn{T^c=\{p\in\RR^3:F^c(p)=1/2\}.}
As a quick recap, SNARF \cite{chen2021snarf} jointly optimizes the pose-dependent canonical-space occupancy field $f_{\sigma_f}(\ \cdot\ ,\theta^i):\RR^3\to[0,1]$ and the forward skinning field $w_{\sigma_w}(\ \cdot\ ):\RR^3\to\RR^{24}$, both represented by neural networks with $\sigma_f$ and $\sigma_w$ as the parameters, by minimizing the binary cross entropy (BCE) loss on the predicted occupancy $f_{\sigma_f}(p,\theta_i)$ and the ground truth occupancy label $o(q^i)$ at the warped location $q^i=W(p,w_{\sigma_w}(p),T,\theta^i)$, i.e.,
\eqn{\min_{\sigma_f,\sigma_w}\mcl L_{\rm BCE}(f_{\sigma_f}(p,\theta^i),o(q^i)).\label{eq:snarfobjori}}
However, we empirically found that jointly optimizing $\sigma_f$ and $\sigma_w$ leads to local minima due to the non-uniqueness of solution of \eqref{eq:snarfobjori}. More specifically, an incorrect canonical shape with an incorrect skinning field can accidentally be warped to a correct posed shape (see Fig.~\ref{fig:skinning_test}).

\begin{figure}[t]
\centering
\includegraphics[width=0.95\linewidth]{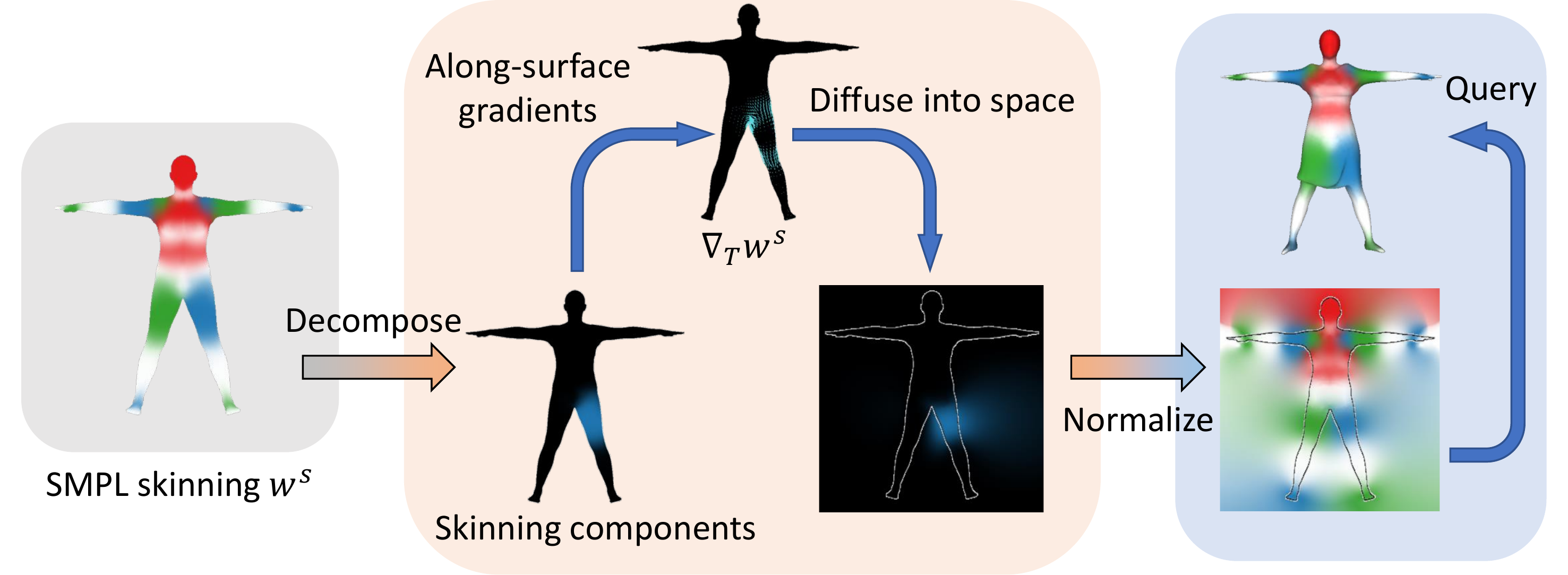}
\caption{Diffused skinning visualized. Each component of the skinning weights on SMPL~\cite{loper2015smpl} is diffused independently and re-normalized to form a skinning field.}
\label{fig:skinning_vis}
\end{figure}

To address this ambiguity, we propose to fix the skinning weights in \eqref{eq:snarfobjori} and let the optimization focus on the occupancy only. This effectively reduces learnable parameters and generates more stable results, which is essential for stage two. A good forward skinning weight field $w:\RR^3\to\RR^{24}$ should satisfy the following constraints: (i) $w(p)$ should be identical to the SMPL skinning weights $w^s(p)$ for $p$ lying on the SMPL body surface $T$; (ii) $w$ should naturally diffuse from the SMPL surface, in the sense that its rate of change along the normal direction should be zero. These lead to the following constraints (equations below should be regarded as component-wise):
\eqn{w(p)=w^s(p),\quad\nabla_p w(p)\cdot n^s(p)=0,\quad\text{for $p\in T$,}\label{eq:diffuseconstraints}}
where $w^s(p)$ denotes the SMPL skinning weights at $p$, and $n^s(p)$ denotes the normal direction of $T$ at $p$. We remark that the gradient of a scalar function on a curved surface is a defined concept and is in fact a tangent field of the surface. Considering each component of $w^s$ as a scalar function on $T$, we compute their gradients along $T$ as $\nabla_Tw^s$. Note that Eq. \eqref{eq:diffuseconstraints} says $\nabla_pw$ is tangential to $T$, and that $w=w^s$ on $T$. Hence, we can equivalently rewrite Eq. \eqref{eq:diffuseconstraints} as
\eqn{w(p)=w^s(p),\quad\nabla_p w(p)=\nabla_Tw^s(p),\quad\text{for $p\in T$}\label{eq:gradconstraintseqver},}
which can be reformulated as minimizing the following energy (with smoothness regularization term $\norm{\nabla^2 w}^2$):
\eqn{\lambda_{\rm p}^s\int_{p\in T}\norm{w(p)-w^s(p)}^2+\lambda_{\rm g}^s\int_{p\in T}\norm{\nabla_pw(p)-\nabla_{T}w^s(p)}^2+\lambda_{\rm reg}^s\int_{\RR^3}\norm{\nabla^2 w}^2,\label{eq:weightoptobj}}
where the $\lambda$'s are weights to ensure numerical stability. We apply an off-the-shelf solver \cite{kazhdanpointinterpolant} to obtain $w$. Note that each component of $w$ is solved separately, clamped to the range of $[0,1]$ and finally re-normalized to sum up to $1$ (Fig.~\ref{fig:skinning_vis}). Please refer to the supplementary material for more details.

Having solved $w$ from Eq. \eqref{eq:weightoptobj}, we fix it in SNARF \cite{chen2021snarf} and train $F_{\sigma_f}$ in \eqref{eq:snarfobjori}. Techniques such as multiple correspondences in are still employed (see the supplementary material for details). Stage one terminates as soon as a coarse shape is available, which is much shorter than the original setup which strives for fine details. After training, we set $F^c$ as:
\eqn{F^c(p)=F_{\sigma_f}(p,\theta^{i_0}),\quad\text{where $\theta^{i_0}=\arg\min\{\norm{\theta^i}_1:\theta_i$ in the training poses$\}$}.}
In other words, we pick the canonical shape closest to the zero-pose in $L_1$-norm out of all training poses. Note that although $F^c$ is not pose-dependent, it is enough for our purpose. Finally, we extract the $1/2$-level-set $T^c$ of $F^c(p)$ as our canonical template, which can be posed via LBS with skinning weights queried from $w$. This concludes stage one.

\subsection{Stage Two: Modeling Pose-Dependent Clothing Deformations}\label{sec:stage2}

After obtaining the canonical template $T^c$ representing the coarse clothing topology in stage one, we further predict pose-dependent offsets from its surface. First, we uniformly sample a point set $\{p^c_k\}_{k=1}^{N_c}$ on the learned canonical template surface $T^c$ and query their skinning weights in the diffused skinning field: $w(p^c_k)$. Moreover, we follow \cite{ma2021pop} to assign a geometric feature vector $g_k\in\RR^{C_{\rm geom}}$ to each $p^c_k$, learned in an auto-decoding fashion~\cite{park2019deepsdf}. For a specific pose $\theta$, the final output point set $\{q_k\}_{k=1}^{N_c}$ representing pose-dependent deformations is obtained by offsetting the canonical point set after applying LBS warping. In addition, we take into consideration the possible inaccuracy in $T^c$ and add a template correction offset $c_k$, leading to the formulation
\eqn{q_k=W(p^c_k+c_k,w(p^c_k),T,\theta)+r_k.\label{eq:basepluscorplusres}}
Since $c_k$ are corrections made to the template itself, they are designed to be pose-agnostic. In the rest of this section, we introduce how to obtain $c_k$ and $r_k$.

\subsubsection{Pose-Agnostic Template Correction}

Since $T^c$ is only coarsely trained, it may have not fully converged to align with training scans. For example, $T^c$ may lack facial details, which is generally pose-{\it in}dependent, but requiring only $r_k$ in Eq.~\eqref{eq:basepluscorplusres} to account for both pose-dependent deformation and pose-independent correction leads to sub-optimal performance. We propose a template correction offset $c_k$ that is pose-agnostic, obtained by feeding the geometric feature $g_k$ to a 4-layer MLP $C(\ \cdot\ )$. The offset $c_k$ is added to $p_k^c$ before LBS warping.

\subsubsection{Projection-Based Pose Encoding}

To generate pose-dependent offsets $r_k$ in Eq.~\eqref{eq:basepluscorplusres}, we need to encode pose-dependent features for $r_k$ to condition on. POP~\cite{ma2021pop}, which greatly inspired our work, render the coordinates of posed vertices to the UV-space, and encode them with U-Nets~\cite{ronneberger2015unet}. This scheme provides a continuous feature space over the SMPL body surface. However, the continuity of UV-space features only extends up to the boundaries of the UV islands. Moreover, in our case there is no predefined UV mapping for our templates.

To adapt such pose encoding to a more general setting where templates do not have predefined UV, and to make the feature space more continuous, we propose to directly render the posed coordinates to images instead of UV maps. First, we extract the template surface $T^c$ as a triangle mesh $M^c$, with vertices $\{v^c_k\}$ and associated skinning weights $\{w(v^c_k)\}$. We then warp $v^c_k$ to the $i$-th pose: $u^i_k=W(v^c_k,w(v^c_k),T,\theta_i)$. Next, we color the mesh $M^c$ by attaching the coordinates of $u^i_k$ to the vertex $v^c_k$ as its ``color''. Finally we render the ``colored'' mesh $M^c$ to images with orthographic projections. Each pixel of the rendered images contains the coordinates of the corresponding {\it posed} vertices. We also adopt a multi-view setup for better coverage of the surface. We choose $N_{\rm v}=4$ views, looking at the template from its left-front side, left-back side, right-front side and right-back side. Moreover, each view is slightly tilted to provide coverage for the top of the head and the bottom of the feet. Following \cite{ma2021pop}, we refer to these rendered images as position maps.

Let the position maps for the $i$-th pose be denoted as $I^i_{d}\in\RR^{H\x W\x3}$, where $d=1,2,3,4$ is the index for 4 viewing directions. We feed them to U-Net \cite{ronneberger2015unet} encoders $U_d$ (one for each view, but shared across all poses) to extract pose-dependent features $z^i_{d}=U_d(I^i_{d})\in\RR^{H\x W\x C_{\rm pose}}$, where $C_{\rm pose}$ is the number of channels of feature maps. With encoded position maps, we are able to extract pixel-aligned features for an arbitrary point $p$ on $T^c$ by first projecting it to each image and then querying the pixel feature via bilinear interpolation. The sampled pixel-aligned feature is denoted as $z^i_{d}(p)\in\RR^{C_{\rm pose}}$. We concatenate the sampled features from all views as the final pose feature for $p$, denoted as $z^i(p)=[z^i_1(p),z^i_2(p),z^i_3(p),z^i_4(p)]\in\RR^{N_{\rm v}\cdot C_{\rm pose}}$, where $[\ \cdots\ ]$ denotes concatenation.

\subsubsection{Decoding Pose-Dependent Deformations}

The final step is to generate $r_k$ in Eq.~\eqref{eq:basepluscorplusres} and associated normals $n_k$ conditioned on projection-based pose features to represent fine details and pose-dependent clothing deformations. Following \cite{ma2021pop}, we decode the $r_k$ and $n_k$ with an 8-layer MLP $D(\ \cdot\ )$:
\eqna{
    [r^c_k,n^c_k]&=&D([z^{i}(p^c_k),g_k]),\quad r^c_k,n^c_k\in\RR^3.
}
In addition, we also employ local transformations as in \cite{ma2021pop}. Recall that $R^i_j$ in Eq.~\eqref{eq:warpingeq} are rigid transformations determined by $\theta^i$. Let $\hat R^i_j$ be the rotation part of $R^i_j$. Then we apply the weighted combination of $\hat R^i_j$ to the output of the decoder $D(\ \cdot\ )$, i.e., $r_k=\sum_{j=1}^{24}w_j(p^c_k)\hat R^i_j(r^c_k)$ and $n_k=n_k'/\norm{n_k'}$, where $n_k'=\sum_{j=1}^{24}w_j(p^c_k)\hat R^i_j(n^c_k)$. Plugging $r_k$ into \eqref{eq:basepluscorplusres}, together with normals $n_k$, gives the final point cloud (with normals) $\{(q_k,n_k)\}_{k=1}^{N_c}$ of our method.

When multiple outfits are present in the input data, the templates for different outfits are trained separately in stage one, but share the template corrector $C$, the pose encoders $U_d$, and the deformation decoder $D$. By sharing the neural networks in stage two for all outfits, they can learn clothing deformation patterns in common for different outfit styles.  

\subsection{Training Losses}

For stage one, we did not modify the training loss and training procedure of SNARF~\cite{chen2021snarf}. Interested readers are referred to the original paper for more details. For stage two, we define the following loss terms in the spirit of \cite{ma2021pop}:
\eqn{\mcl L_{\rm total}=\lambda_{\rm p}\mcl L_{\rm p}+\lambda_{\rm n}\mcl L_{\rm n}+\lambda_{\rm c,reg}\mcl L_{\rm c,reg}+\lambda_{\rm r,reg}\mcl L_{\rm r,reg}+\lambda_{\rm g,reg}\mcl L_{\rm g,reg},\label{eq:loss_total}}
where the $\lambda$'s are the weighting coefficient for different loss terms. The first two terms $\mcl L_{\rm p}$ and $\mcl L_{\rm n}$ are losses on the point cloud and the normals, respectively. More specifically, let $P^{\rm gt}=\{(p^{\rm gt}_k,n^{\rm gt}_k)\}_{k=1}^{N_{\rm gt}}$ be the ground truth point cloud and let $P^{\rm pd}=\{(q^{\rm pd}_k,n^{\rm pd}_k)\}_{k=1}^{N_{\rm pd}}$ be the predicted point cloud (with normals). Then
\eqna{
    \mcl L_{\rm p} &=& \frac{1}{N_{\rm pd}}\sum_{k=1}^{N_{\rm pd}}\min_{k'}\prn{(q_k^{\rm pd}-p_{k'}^{\rm gt})\cdot n^{\rm gt}_{k'}}^2+\frac{1}{N_{\rm gt}}\sum_{k'=1}^{N_{\rm gt}}\min_{k}\norm{p_{k'}^{\rm gt}-q_k^{\rm pd}}^2_2,\\
    \mcl L_{\rm n} &=& \frac{1}{N_{\rm pd}}\sum_{k=1}^{N_{\rm pd}}\norm{n^{\rm pd}_k-n^{\rm gt}_{k'}}_1,\text{ where }k'=\mbox{argmin}_{k'}\norm{q^{\rm pd}_k-p^{\rm gt}_{k'}}.
}
The other three terms are regularization terms on the template correction offsets, the pose-dependent offsets, and the per-point geometric features, respectively:
\eqn{
    \mcl L_{\rm c,reg}=\frac{1}{N_{\rm pd}}\sum_{k=1}^{N_{\rm pd}}\norm{c_k}_2^2,\ \mcl L_{\rm r,reg}=\frac{1}{N_{\rm pd}}\sum_{k=1}^{N_{\rm pd}}\norm{r_k}_2^2,\ \mcl L_{\rm g,reg}=\frac{1}{N_{\rm pd}}\sum_{k=1}^{N_{\rm pd}}\norm{g_k}_2^2.
}
Please refer to the supplementary material for details of model architectures, hyper-parameter settings and the training procedure.

\section{Experiments}

In this section we evaluate the representation ability of our method. Due to space limits, we report the results for interpolation, extrapolation and novel scan animation here, and provide extended evaluations, ablation studies and failure cases in the supplementary material.

\subsection{Evaluation Details}

\subsubsection{Baselines}

We compare our method with the SOTA clothed human modeling methods: POP~\cite{ma2021pop}, SNARF~\cite{chen2021snarf}, and SCANimate~\cite{saito2021scanimate}. POP \cite{ma2021pop} also adopts a point cloud representation which is conditioned on SMPL/SMPL-X~\cite{loper2015smpl,pavlakos2019smplx} templates. SNARF~\cite{chen2021snarf} and SCANimate~\cite{saito2021scanimate} are currently the SOTA implicit methods with learned volumetric skinning fields.

\subsubsection{Datasets}

We evaluate our method on two large-scale datasets with multiple outfits: ReSynth~\cite{ma2021pop} and CAPE~\cite{ma2020cape}. We follow POP~\cite{ma2021pop} to use 12 outfits and 14 outfits from ReSynth and CAPE, respectively, for cross-outfit training. Note that ReSynth is much more diverse in outfit types than CAPE, and serves as a more convincible test for modeling outfits with different topologies. Moreover, since the implicit baselines do not output point clouds, and the generated point sets from FITE and POP have different densities, we use Screened Poisson Reconstruction~\cite{kazhdan2013spr} to obtain closed meshes for evaluation. Please refer to the supplementary material for more details on dataset preprocessing.

\subsubsection{Metrics}

Following the common evaluation pipeline~\cite{ma2021pop}, we use the Chamfer-$L_2$ distance $d_{\rm cham.}$ (lower is better) and cosine similarity $S_{\rm cos}$~\cite{park2019deepsdf} (higher is better) to measure the error of generated clothed humans. Due to the stochastic nature of clothing, error measurement with the ground truth scans in extrapolated poses does not faithfully reflect the modeling quality. Thus, following previous work~\cite{ma2021pop,saito2021scanimate}, we conduct a large-scale user study to evaluate the visual quality of different methods for extrapolation experiments. During the user study, each viewer is given either a pair of point clouds or a pair of meshes placed side-by-side, and is asked to vote on the one with higher \tf{overall} visual quality after considering factors such as realism, details and artifacts. The left-right order is randomly shuffled to prevent the preference for a certain side. The choice of presented outfit and pose is also random with equal probability.

\subsection{Interpolation Experiments}

We evaluate the representation ability of our method with interpolation experiments on the ReSynth dataset and the CAPE dataset. Considering dataset sizes, for training, we choose every 2nd frame for ReSynth and choose every 4th frame for CAPE, both from their official training splits. The rest of the training sequences are used for evaluation.

Table~\ref{table:exp_interpolation} shows the quantitative results of the interpolation evaluation. Note that the modeling difficulty, as well as the error distribution, varies drastically from outfit to outfit. We thus report the quantitative results for each outfit separately. Due to page limits, we report three outfits from CAPE and three from ReSynth and present more in the supplementary material. The results in Table~\ref{table:exp_interpolation} shows that both point-based methods, FITE and POP, outperform implicit methods by a large margin. Between FITE and POP, our method performs notably better for outfits that greatly differ from the minimal body (long dress). Note that the benefit of cross-outfit training for FITE can be more clearly observed from the long dress example. This is due to the fact that projection-based encoding is harder to train than UV encoding and can thus benefit more from the regularization effect brought by cross-outfit training. We will discuss this more closely in the supplementary material.

\setlength{\tabcolsep}{1.0pt}
\begin{table}
\scriptsize
\begin{center}
\caption{Quantitative results of interpolation experiments. Since SNARF~\cite{chen2021snarf} and SCANimate~\cite{saito2021scanimate} do not support cross-outfit modeling, we evaluate the outfit-specific versions of POP and FITE for fairness (denoted as POP-OS and FITE-OS). Note that $d_{\rm cham}$ reported below have been multiplied by $10^5$.}
\label{table:exp_interpolation}
\begin{tabular}{l|cccccc|cccccc}
\hline
\multirow{3}{*}{Method} & \multicolumn{6}{c|}{CAPE Data} & \multicolumn{6}{c}{ReSynth Data} \\
\cline{2-13}
 & \multicolumn{2}{c}{00096} & \multicolumn{2}{c}{00215} & \multicolumn{2}{c|}{03375}& \multicolumn{2}{c}{Carla 004} & \multicolumn{2}{c}{Christine 027} & \multicolumn{2}{c}{Felice 004} \\
  & \multicolumn{2}{c}{jerseyshort} & \multicolumn{2}{c}{poloshort} & \multicolumn{2}{c|}{blazerlong}& \multicolumn{2}{c}{long pants} & \multicolumn{2}{c}{short dress} & \multicolumn{2}{c}{long dress} \\
 & $d_{\rm cham}$ & $S_{\rm cos}$ & $d_{\rm cham}$ & $S_{\rm cos}$ & $d_{\rm cham}$ & $S_{\rm cos}$  & $d_{\rm cham}$ & $S_{\rm cos}$ & $d_{\rm cham}$ & $S_{\rm cos}$ & $d_{\rm cham}$ & $S_{\rm cos}$ \\
\hline
SCANimate~\cite{saito2021scanimate} & 0.632 & 0.942 & 0.730 & 0.927 & 0.957 & 0.914  & 0.721 & 0.943 & 1.750 & 0.940 & 17.578 & 0.803\\
SNARF~\cite{chen2021snarf}& 0.155 & 0.964 & 0.191 & 0.941 & 0.624 & 0.929 & 0.340 & 0.949 & 0.621 & 0.953& 2.426 & 0.906 \\
POP-OS~\cite{ma2021pop}& \tf{0.036} & \tf{0.987} & 0.084 & 0.980 & \tf{0.249} & \tf{0.967}  & 0.507 & 0.940 & 0.437 & \tf{0.964} & 1.591 & 0.925 \\
POP~\cite{ma2021pop}  & 0.044 & 0.986 & 0.091 & 0.978 & 0.252 & 0.966& 0.485 & 0.939 & \tf{0.421} & 0.960 & 1.718 & 0.920\\
\hline
FITE-OS & 0.041 & \tf{0.987} & \tf{0.074} & \tf{0.981} & 0.271 & 0.966  & 0.300 & \tf{0.957} & 0.462 & \tf{0.964} & 1.805 & 0.918  \\
FITE & 0.042 & \tf{0.987} & 0.076 & 0.980 & 0.274 & 0.964  & \tf{0.299} & 0.956 & 0.455 & 0.963 & \tf{1.355} & \tf{0.933} \\
\hline
& \multicolumn{2}{l}{Tight clothing} & \multicolumn{8}{c}{$\Longrightarrow$} & \multicolumn{2}{r}{Loose clothing}\\
\hline
\end{tabular}
\end{center}
\end{table}
\setlength{\tabcolsep}{1.4pt}

\begin{figure}
\centering
\includegraphics[width=\linewidth]{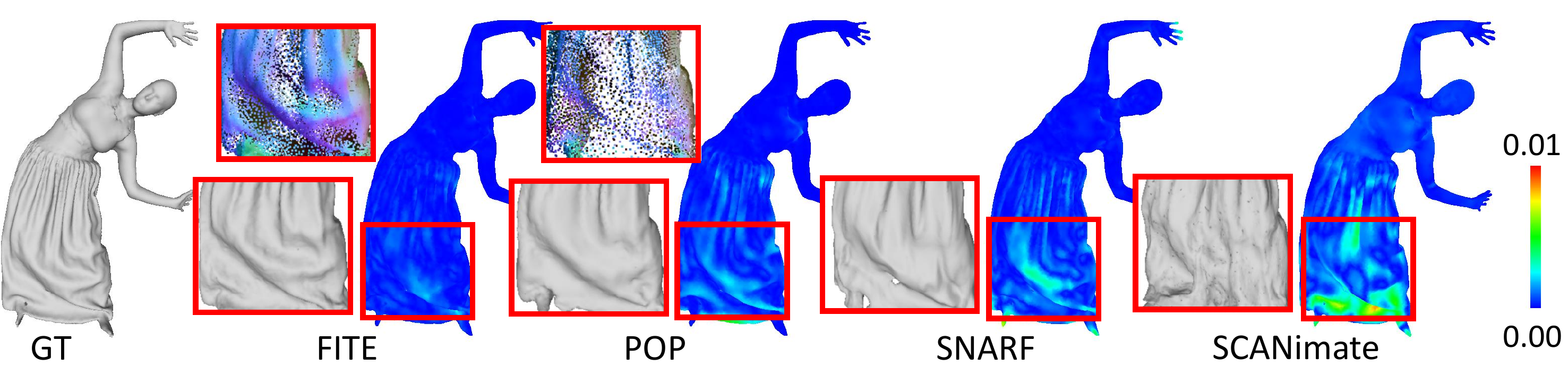}
\caption{Qualitative results of the interpolation experiment. Error maps are visualized w.r.t. the largest error in this comparison.}
\label{fig:exp_interp}
\end{figure}

The improvement can be more obviously observed in Fig.~\ref{fig:exp_interp}. For a long dress, FITE generates densely distributed points for the loose part and is able to represent more details, while the output of POP becomes sparse and can only model a coarse shape. For implicit methods, the modeling of details is less faithful to the ground truth and perceptually less realistic.

\subsection{Extrapolation Experiments}

\begin{figure}
\centering
\includegraphics[width=\linewidth]{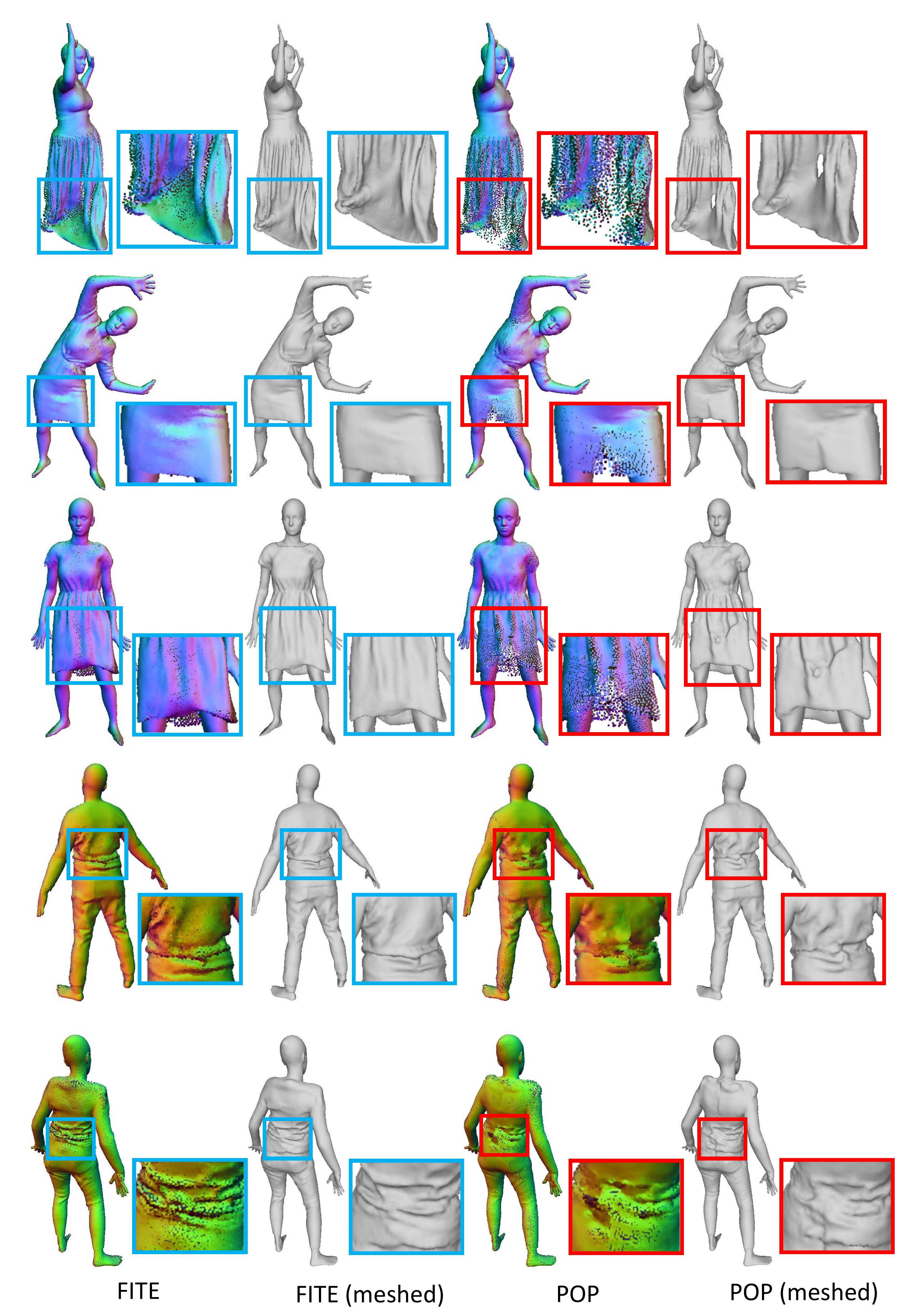}
\caption{Qualitative results of pose extrapolation.}
\label{fig:exp_extrap}
\end{figure}

For extrapolation experiments, we use the official training sequences (full data, multi-outfit) and test sequences. Fig.~\ref{fig:exp_extrap} shows qualitative comparisons of our method and POP, with seen outfits in unseen poses. For long dresses, POP produces overly sparse points and fail to represent the surface. For short dresses, POP must deform points from the legs of the SMPL-X template~\cite{pavlakos2019smplx} to form the dresses. This incoherency leads to the discontinuity on the dresses. Even for tight clothing, the discontinuity in the UV space also leaves visible seams on the clothing. On the other hand, FITE utilizes templates that already capture the clothing topology and encode pose information in a multi-view projection scheme, producing outputs topologically coherent with given data.

We also conduct a large-scale user study to evaluate quantitatively the extrapolation performance (421 participants, each with 20 votes; 8420 votes in total). Among all votes we received, in terms of generated point clouds (4690 votes), 75.42\% prefer FITE over POP (24.58\%); in terms of reconstructed meshes (3730 votes), 59.37\% prefer FITE over POP (40.63\%). Although surface reconstruction partly compensates the drawbacks of POP, the perceptual advantage of our method is still clearly observable. As a final remark, the artifacts of POP in Fig.~\ref{fig:exp_extrap} are not clearly observable in the training set, but they appear frequently in the test set. We believe this reveals that the generalizability of POP is essentially limited by the fixed underlying body template. Please refer to the supplementary material for more details on how the user study is conducted.

\subsection{Novel Scan Animation}
\begin{figure}
\centering
\includegraphics[width=\linewidth]{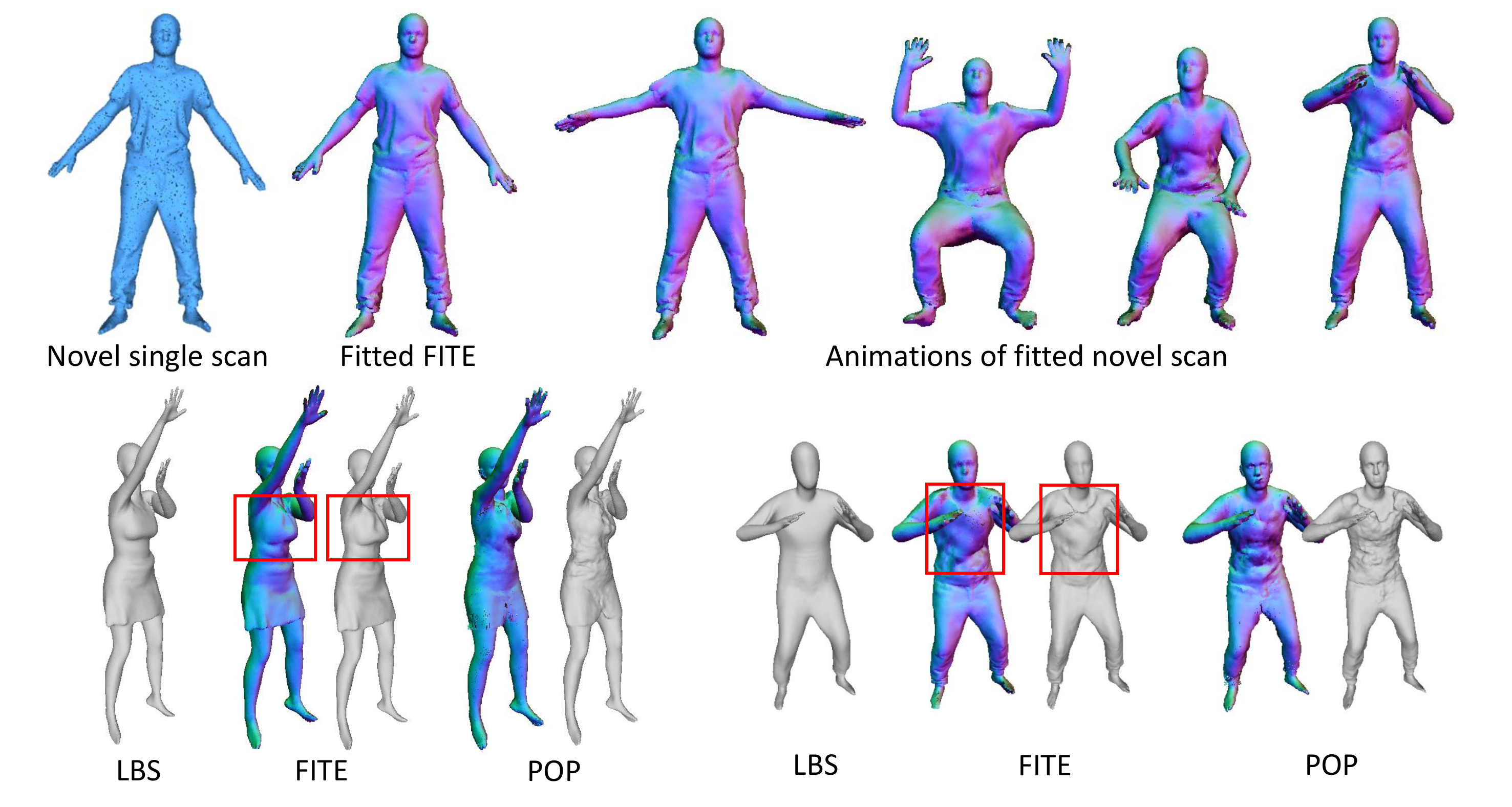}
\caption{Novel scan animation results (pose-dependent offsets highlighted).}
\label{fig:novelfit}
\end{figure}

In stage two, the networks $C,\ U_d$ and $D$ are shared across outfits and learn the common deformation pattern for different outfits. Thus, they can be used to fit novel scans by optimizing the geometric features only w.r.t. a new scan. Fig.~\ref{fig:novelfit} shows the generalization to novel scans. Our method adds pose-dependent to LBS warped templates and produces less noise than POP~\cite{ma2021pop}. Please refer to the supplementary material for more details.

\section{Conclusion}

We present FITE, a first-implicit-then-explicit framework for modeling clothed humans with realistic pose-dependent deformations. Evaluated on outfits with different topologies, our method is shown to outperform previous methods by incorporating the merits of both the implicit representation and the point set representation. Moreover, we believe several individual modules in this framework can also inspire related search, namely, diffused skinning as a smooth interpolation of the learned SMPL skinning weights, and projection-based pose encoding for introducing a continuous feature space on arbitrary mesh surfaces. However, as is currently formulated, several aspects still require further exploration.

\paragraph{Unifying canonical templates} In the current framework, stage one learns the coarse templates for each outfit separately. It is worthwhile to explore unifying different outfits with a single shape network, i.e., learning not only a prior for deformations, but also a prior for the outfits, which can lead to faster and more stable outfit generalization.

\paragraph{Driving the underlying templates} After obtaining coarse templates in stage one, LBS is applied for reposing. However, LBS does not always reflect true clothing motions, especially for loose clothing in extreme poses. Replacing LBS in stage one with coarse-level physics-based simulation can possibly improve the performance for certain outfits.

\paragraph{Disentanglement of clothing and pose} In the second stage of FITE, the projection-based encoders are used to extract pose information from rendered position maps. However, these position maps already contain the clothing information, and thus clothing and pose are not fully disentangled. Future work should explore a representation that further disentangles these factors.\\

\noindent\textbf{Acknowledgements.} This paper is supported by National Key R\&D Program of China (2021ZD0113501) and the NSFC project No.62125107 and No.61827805.

\bibliographystyle{splncs04}
\bibliography{egbib}
\end{document}